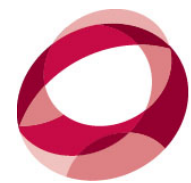

# Advances in Artificial Intelligence Require Progress Across all of Computer Science

February 2017

*Gregory D. Hager, Randal Bryant, Eric Horvitz, Maja Matarić, and Vasant Honavar*

Over the last decade, the constellation of computing technologies referred to as artificial intelligence (AI) has emerged into the public view as an important frontier of technological innovation with potential influences in many realms. Advances in many disciplines related to AI, including machine learning, robotics, computer vision, natural language processing, inference, decision-making, and planning, are contributing to new-fielded products, services, and experiences. Offerings such as navigation systems, web search, speech recognition, machine translation, face recognition, and recommender engines have become part of the daily life of millions of people. Other applications coming to the fore include semi-autonomous and autonomous ground and air vehicles, systems that harness planning and scheduling, intelligent tutoring, robotics. More broadly, cyber-physical and robotic systems, incorporating varying degrees of AI technology, are poised to be fielded in a variety of real-world settings.

Although AI will be an engine for progress in many areas, creating real-world systems that realize these innovations will in fact require significant advances in virtually all areas of computing, including areas that are not traditionally recognized as being important to AI research and development. Indeed, we expect future AI systems will not only draw from methods, tools, and themes in other areas of computer science research, but will also provide new directions for research in areas such as efficiency, trustworthiness, transparency, reliability, and security. This is not new – the history of AI is inextricably intertwined with the history of advances in broader computer science (CS) as well as applications in related areas such as speech and language, computer vision, robotics, and other types of intelligent systems.

In what follows, we review several promising areas of interaction between AI and broader computer science that provide rich opportunities ahead for research and development. We include opportunities that seem especially important as AI systems become more ubiquitous and are playing roles critical to our individual and combined health and well-being. In brief, we see particularly rich opportunities for supporting advances in AI via synergies and collaboration with research and development in the following areas, described briefly below, then further expanded upon:

**Computing systems and hardware.** There are opportunities ahead for leveraging innovations in computing systems and hardware. Directions include the development of methods for speeding up core computational procedures employed in AI systems, such as the methods used to train and to execute classification for perceptual tasks using convolutional neural networks. Opportunities include new approaches to parallelism, smart caching, and uses of specialized hardware like FPGAs to lower costs of computation and to meet the demands and robustness needed with AI applications.

**Theoretical computer science.** AI was built on theoretical work based on the mathematics of computability in the early 20th century by Turing, Church, and others. AI challenges have long posed and faced combinatorial challenges and has made use of results on the performance and precision of approximation procedures. There are continuing opportunities for the formal study of hard challenges in AI with tools and techniques developed in the realms of analysis of algorithms, including efforts in combinatorics, computational complexity theory, and studies of computability.

**Cybersecurity.** AI systems are being developed for high-stakes systems in such areas as healthcare and transportation. These systems are also bringing to the fore new attack surfaces that need to be understood and protected. Directions include understanding and hardening systems to whole new categories of attack including, "machine learning attacks," where clever adversarial procedures are employed to inject data into systems that will confuse or bias them in their intended operation. AI systems frame new challenges that will require advances in security that address the new attack surfaces to ensure that they are safe, reliable, robust and secure against malicious attacks.

**Formal methods**. Formal methods can play a critical role in defining and constraining AI systems, so as to ensure that their behavior conforms to specifications. Efforts include methods for doing formal verification of programs and also to perform real-time verification of systems through new kinds of monitoring. Formal methods are promising approaches to ensuring that programs do not take actions beyond specified goals and constraints.

**Programming languages, tools, and environments.** New programming languages, tools, and programming environments can help engineers to build, test, and refine AI systems. Higher-level languages can offer engineers and scientists new kinds of abstractions and power to weave together multiple competencies, such as a vision, speech recognition, and natural language understanding so as to be able to develop and debug programs that rely on the close coordination of multiple AI analytical pipelines.

**Human-computer interaction**. The key challenges with AI frame numerous opportunities in the broad realm of research in human-computer interaction (HCI), an important area of computer science research. Efforts include methods for explaining the results of AI systems to people, allowing people to work interactively with AI systems (e.g., interactive machine learning), that help with the specification, encoding, and understanding of the implications of different policies, values, and preferences assumed by automated systems, and supporting new kinds of human-AI collaboration, including mixed-initiative interaction and augmenting human cognition.

**Computing Systems and Hardware**

Research in computer hardware, operating systems, and computer networking have been, and continue to be, critical to creating the large-scale systems, including AI systems that use vast amounts of data to build predictive models, optimize complex objective functions, perform automated inference over large knowledge bases, or complex probability distributions. Some of the largest computer facilities in the world are operated by industrial and government organizations to collect data, build predictive models from data using machine learning algorithms, and then use the resulting predictive models to provide services ranging from driving directions to shopping recommendations.

As AI continues to grow, the demands on these systems will also grow and change. For example, current uses of cloud computing are typically not hard real-time. Whether a search query is answered in 0.2s or 0.4s may not matter much in practice. However, a query from a car moving at 60mph may need to be answered within a hard real-time constraint to ensure the safe operation of the automobile.

Historically, special purpose hardware architectures for specific tasks e.g., computer vision, have been in and out of fashion. However, as we approach the end of the Moore's law, performance gains that are required for successful deployment of AI systems in real-world applications are likely to require innovations in hardware and systems as well as co-design of hardware and software to exploit the special features of specialized architectures and systems. Large server systems already use GPUs extensively to support modern learning algorithms. At scale, it may become cost and energy efficient to include even more customized capability that must be shared among one to a million users depending on the system in question.

Unlike many computing tasks, machine learning can often make use of approximate results. They make decisions using statistical models and there can be a tolerance for small computational errors introduced by the hardware or by imperfect synchronization across computing resources. This opens up many possibilities for making the system more energy efficient or scalable at the hardware, software, and system levels. Achieving these benefits will require a careful characterization of the nature of the computational errors that can be introduced and their impact on the overall system functionality.

**Theoretical CS: Analysis of Algorithms, Combinatorics, and Complexity**

AI has influenced, and benefited from, advances in algorithms in a number of areas including automated reasoning, search, planning, optimization, and learning. For example, machine learning, which lies at the heart of many modern AI systems, enables Google's AlphaGo program to devise strategies for playing the game of Go by analyzing millions of moves made by humans in Go tournaments; Amazon's recommender systems to analyze large data sets of transactions and then suggest products to customers; automobile lane tracking systems to detect lane markers from video images and warn drivers when they are veering out of their lanes; language translation systems to generate mappings from one language to another by processing large collections of human-generated translations. Machine learning requires algorithms that construct sophisticated predictive models to offer useful insights and actionable knowledge from large and complex data sets. The demands of machine learning has led to advances in algorithms, especially for optimization of complex objective functions, reasoning about complex probability distributions e.g., using factorized representations, etc. With the emergence of big data, the requirement that machine learning algorithms need to be scalable to massive amounts of high dimensional data, robust in the presence of noise, etc., present many challenges in design and analysis of algorithms.

Similarly, AI systems for planning (e.g., motion planning, dialog planning, and activity planning) call for advances in data structures and algorithms for representing and reasoning about large state spaces, coping with uncertainty, and providing compact representations that support efficient access to data. The emergence of large knowledge bases that codify human knowledge in a variety of domains calls for robust algorithms for automated inference; for example, for answering complex queries against knowledge bases. Many problems, including assembling products that meet specified requirements from available components, can be formulated as problems in search or optimization, and hence, can drive advances in constraint processing and optimization. AI systems for computer vision, natural language dialogue, text processing, video analytics, and other tasks can similarly drive innovations in algorithms.

The success of AI in reducing more and more tasks that were thought to require human intelligence into ones that can be solved algorithmically will stimulate work on establishing the correctness, performance bounds, and the trade-offs among them. This will have immediate practical consequences, as will negative results that establish the theoretical limits of algorithms. Such investigations can lead to important insights, for example whether a task is learnable in principle, in practice, and under what conditions, such as, from observations, queries, or experimentation. Ensuring robust operation will require finding bounds on the potential deviations of an AI system when presented with unanticipated inputs.

**Cybersecurity**

In addition to the many benefits computer technology and the Internet provide society, it has proved to be a powerful tool for adversaries, including malicious individuals, criminals, and nation states. Attackers use a number of methods to extract sensitive information from organizations and to disable or disrupt the activities of individuals, corporations, and government.

AI systems have vulnerabilities that include those of both traditional and new computer systems. For example, by corrupting the training data, an AI system can be tricked into constructing an invalid predictive model. Given the limited mechanisms currently available for testing these models such corruption may be difficult to detect.

As AI systems are deployed in the real world, adversaries will seek ways to trick them into behaving in undesirable ways. For example, imagine attempting to deploy an autonomous armored truck. Adversaries would be highly motivated to force it to stop or to alter its course by putting the truck's control system outside the range of conditions it has been trained to consider. Indeed, such limitations will impede the deployment of AI systems in security-critical environments.

AI can also provide a powerful tool for both cyberattackers and cyberdefenders. On the attack side, a network of botnets could use real-time data analytics to dynamically adapt their behavior, increasing their effectiveness and diminishing their detectability. Conversely, machine learning is already providing an effective tool for detecting anomalies in computer systems that could indicate an intrusion.

Hence, increasing reliance on AI systems as integral components of complex systems, calls for advances in cybersecurity.

**Formal Verification**

Computer systems often fail due to errors in the design or implementation of its hardware and software. Sometimes, such failures, such as the loss of an important file, while annoying, are of no major consequence. But when computers directly control critical systems, such as medical devices, civil infrastructure, and defense systems, consequences of failures are much more severe. Hence, we need tools for ensuring that software, particularly, complex AI software, comply with specifications.

Formal verification tools provide ways to test the correctness and hardware and software under all possible operating conditions. They go beyond the traditional method of testing the system on many different individual cases to consider all possible cases. Such tools have had significant success with purely digital computations. For example, Intel took a loss of $475 million when the first version of its Pentium processor could, under very rare circumstances, produce the wrong result when dividing two numbers. They subsequently developed a set of tools to ensure the correctness of their arithmetic circuits for all possible data. Similarly, other companies have developed and deployed tools that can detect many classes of software errors.

However, establishing the correctness of computer systems that operate in the physical world – either systems whose performance depends on the data that they were trained on, as in the case of AI systems that rely on machine learning, or systems that need to operate in open environments that are hard-to-characterize (e.g., household robots, automated vehicles, etc.) – is much more difficult. For example, it is impossible to anticipate all of the situations an autonomous vehicle might encounter, much less guarantee that it will handle each such situation correctly. On the other hand, it may be possible to use formal verification techniques to verify that AI systems will conform to specific bounds on their performance in such environments.

**Programming Languages**

Rapid development of complex AI systems calls for advances in programming languages and tools. Of particular interest are domain-specific programming languages with built-in high level abstractions that make it easy to design and program AI systems for classes of applications e.g., natural language processing, computer vision, multi-robot systems. Also of interest are new programming languages that support probabilistic computations, large-scale automated inference, constraint processing., etc.

Machine learning also provides an inspiration for a new conception of programming. With these systems, the human programmers create a framework for how the system should operate, but the actual "program", such as the weights that specify the detailed configuration of a neural network, is derived algorithmically by training the system over large amounts of data. One can imagine a future in which even higher levels of automation are applied to construct much of what is done by human programmers today.

**Human-Computer Interaction**

AI will revolutionize the type of research that can and must be done to enable people to effectively use, understand, interact and co-exist with AI systems. To pick one example, today most voice queries are either dictation of a message or order, or a one-time query. Future AI systems will need to engage in broader forms of dialog to deal with ambiguity, confusion, or to improve engagement. With the increasing adoption of AI systems (e.g., automated vehicles, robots, software assistants, in real-world environments) there is an increasing need for research on frameworks, languages, abstraction, and interfaces that allow effective communication and interaction between humans and AI systems.

Of particular interest are mixed initiative systems that allow productive collaborations among AI systems, humans and AI systems, as well as among humans, with mediation and facilitation by humans or AI systems as needed. AI systems must engage with humans in a collaborative manner, allowing people to work interactively with AI systems (e.g., interactive machine learning), that help with the specification, encoding, and understanding of the implications of different policies, values, and preferences assumed by automated systems, and supporting new kinds of human-AI collaboration, including mixed-initiative interaction and augmenting human cognition.

Thus, as capabilities in AI grow, so too will the questions and opportunities to connect AI to people in a meaningful and effective way.

**In Conclusion**

There is a growing and compelling imperative to leverage the advances in AI and automation to improve human lives in many ways. At the same time, such systems will also become far more present and consequential to everyday life, and will provide services and capabilities that will exploit large amounts of data (including personal data), control physical devices of various kinds, including devices in safety critical areas, and be empowered to make and act on decisions of varying importance that could influence individuals and societies in explicit and implicit ways.

The path toward a balanced portfolio of capable, safe, and transparent AI-based systems will draw on a broad spectrum of computing ideas and principles, and is likely to become a driver for new advances in computing. By embracing the promise of AI, we believe that many areas of computer science will not only be advanced, but will also allow AI to address important opportunities and do so in a way that is safe, reliable, and effective.

*This material is based upon work supported by the National Science Foundation under Grant No. 1136993. Any opinions, findings, and conclusions or recommendations expressed in this material are those of the authors and do not necessarily reflect the views of the National Science Foundation.*